\documentclass[letterpaper, 10 pt, conference]{format/ieeeconf}  

\IEEEoverridecommandlockouts                              

\usepackage{dsfont}

\usepackage{amsmath}

\usepackage{float}

\usepackage{graphics} 
\usepackage{graphicx}
\usepackage{booktabs}
\usepackage{array} 
\usepackage{amsmath} 
\usepackage{amssymb}  
\usepackage{todonotes}

\usepackage{amsfonts}
\usepackage{mathtools}
\usepackage[linesnumbered,algoruled,boxed,vlined, noend]{algorithm2e}
\usepackage{amsthm}
\usepackage{comment}
\usepackage{cite}
\usepackage{wrapfig}
\let\labelindent\relax
\usepackage{enumitem}
\usepackage{overpic}
\usepackage{xspace}
\usepackage{multirow}
\usepackage{authblk}

\DeclareMathAlphabet{\mathcal}{OMS}{cmsy}{m}{n} 

\theoremstyle{definition}

\theoremstyle{remark}

\DeclarePairedDelimiterX{\norm}[1]{\lVert}{\rVert}{#1}

{\end{list}}

\usepackage{soul} 
\usepackage[a-2b,mathxmp]{pdfx}[2018/12/22]

\def\ours{MuST\xspace}

\def\progss{ProGSS\xspace}

\newif\ifarxiv
\arxivfalse

\font\titlefont=ptmb at 15.95pt
\title{\titlefont
MuST: Multi-Head Skill Transformer for Long-Horizon Dexterous Manipulation with Skill Progress}
\author{Kai Gao$^{1,2}$\quad Fan Wang$^{1}$\quad Erica Aduh$^{1}$\quad Dylan Randle$^{1}$\quad Jane Shi$^{1}$
\thanks{$^{1}$Amazon Robotics, MA, USA. Email: {\tt\small { \{kaigaoar, fanwanf, aduheric, dylanran, janeshi\}}@amazon.com}.
}
\thanks{$^{2}$Department of Computer Science, Rutgers University, NJ, USA. Email: {\tt\small { \{kg627\}}@cs.rutgers.edu}. Work done during the internship at Amazon Robotics.
}
}
\begin{document}

\maketitle


\begin{abstract}
Robot picking and packing tasks require dexterous manipulation skills, such as rearranging objects to establish a good grasping pose, or placing and pushing items to achieve tight packing. These tasks are challenging for robots due to the complexity and variability of the required actions. To tackle the difficulty of learning and executing long-horizon tasks, we propose a novel framework called the Multi-Head Skill Transformer (MuST). This model is designed to learn and sequentially chain together multiple motion primitives (skills), enabling robots to perform complex sequences of actions effectively. MuST introduces a ``progress value'' for each skill, guiding the robot on which skill to execute next and ensuring smooth transitions between skills. Additionally, our model is capable of expanding its skill set and managing various sequences of sub-tasks efficiently. Extensive experiments in both simulated and real-world environments demonstrate that MuST significantly enhances the robot's ability to perform long-horizon dexterous manipulation tasks.
\end{abstract}

\section{Introduction}\label{sec:intro}
As robotics technology advances, the deployment of robots in everyday tasks is rapidly transitioning from a conceptual idea to a practical reality. Unlike traditional industrial robots, which are typically designed for repetitive, single-purpose tasks, the next generation of robots is expected to perform complex, dexterous manipulation tasks that require the seamless integration of multiple skills and the execution of diverse sub-tasks.

Recently, advances in policy learning~\cite{chi2023diffusion, zhao2023learning, team2024octo} have shown great promise by effectively learning from human demonstrations to address dexterous manipulation tasks that were notoriously difficult to design and program manually. These methods leverage the richness and variety of human demonstrations, capable of capturing multi-modal data, and many also utilize foundation models to improve generalization and learning efficiency~\cite{open_x_embodiment_rt_x_2023, brohan2023rt1roboticstransformerrealworld, brohan2023rt2visionlanguageactionmodelstransfer, ze20243ddiffusionpolicygeneralizable, yang2024equibotsim3equivariantdiffusionpolicy}. However, despite these advances, most systems are designed and tested to handle specific, single tasks and often fall short when faced with long-horizon tasks that require combining and sequencing multiple skills over extended periods~\cite{mandlekar2021learninggeneralizelonghorizontasks, duan2017oneshotimitationlearning}.

\begin{figure}[t]
    \centering
    \includegraphics[width=0.5\textwidth]{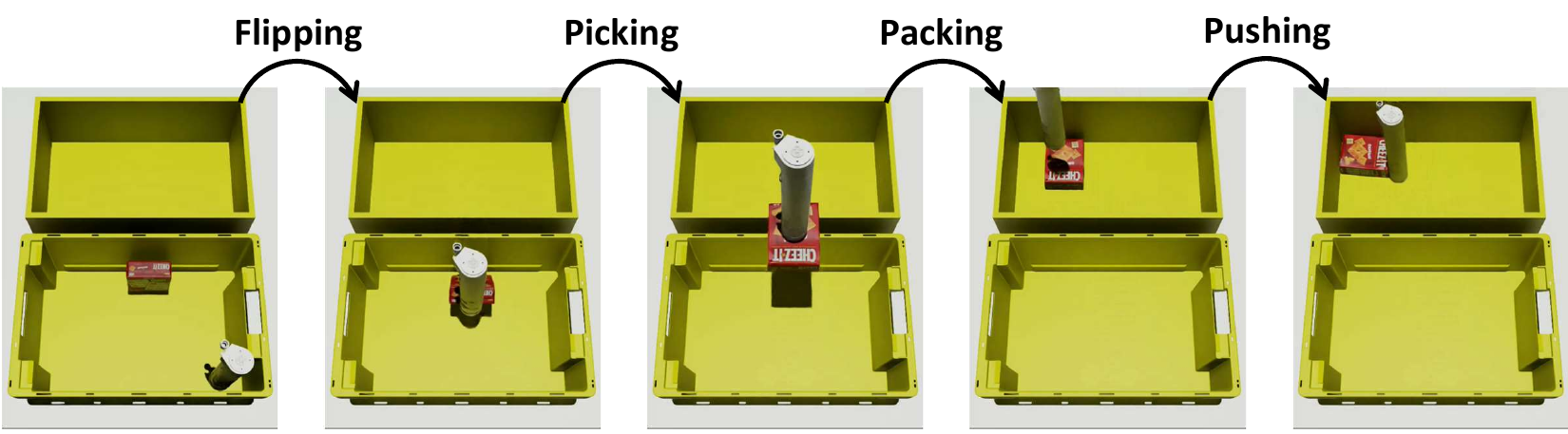}
   \includegraphics[width=0.5\textwidth]{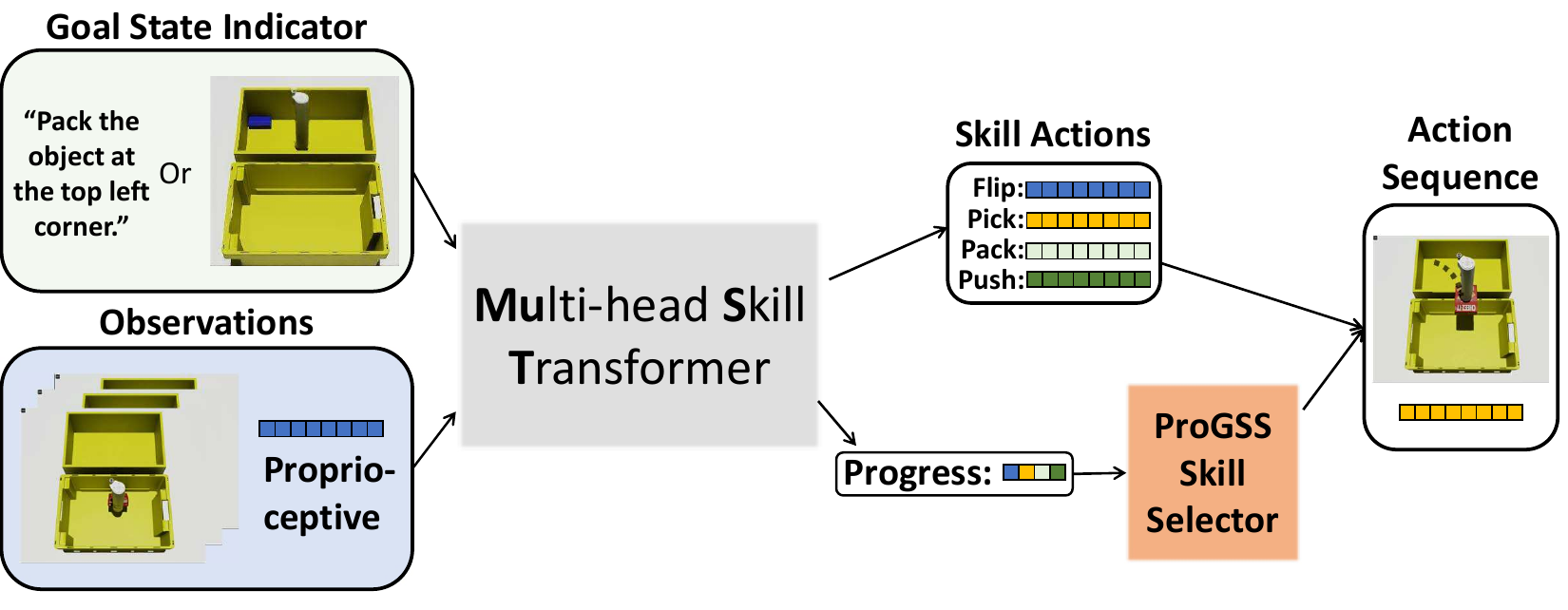}
    \caption{[Top] An example of long-horizon dexterous manipulation. The robot executes four skills to manipulate an object from the boundary of the picking tote to the corner of the packing tote. [Bottom] Our proposed imitation learning model \ours with N-skill and skill selector \progss.}
    \label{fig:intro}
\end{figure}

A key hypothesis we have to improve reliability of policy learning for long-horizon task is that many such tasks can be decomposed into multiple heterogeneous sub-tasks, where each sub-task requires a distinct skill, and these skills are highly reusable. This is particularly evident in warehouse robotics, where tasks such as picking and packing involve dinstinct skills like flipping, grasping, and pushing. For example, Fig.~\ref{fig:intro}[Top] shows a robot flipping an object from the boundary of a picking tote and compactly packing it in the corner of a packing tote. 

Decades of research have explored the problem of chaining multiple skills to achieve long-horizon tasks. Task and Motion Planning (TAMP) methods integrate high-level symbolic reasoning with low-level motion planning, enabling robots to sequence skills while ensuring physical feasibility~\cite{Dantam16, Srivastava14, Wolfe10}. Similarly, reinforcement learning (RL), particularly hierarchical RL, decomposes tasks into sub-skills, facilitating more efficient exploration in complex environments~\cite{kulkarni2016hierarchicaldeepreinforcementlearning, eysenbach2018diversityneedlearningskills, vezhnevets2017feudalnetworkshierarchicalreinforcement, bacon2016optioncriticarchitecture}. However, these methods still face challenges. For instance, RL methods struggle with exploration and scalability, as the search space for multi-skill tasks grows exponentially with complexity~\cite{nasiriany2022augmenting}. 

Instead, we propose a novel approach, \ours (Multi-Head Skill Transformer), designed to enhance the reliability of policy learning by primarily building upon the policy's existing structure, while introducing minimal additional complexity. \ours operates by decomposing long-horizon tasks into sequence of reusable skills. At each timestamp, the appropriate skill is selected based on the current observation and state, enabled by a robust progress estimator for each skill and a skill selector. 

Specifically, \ours extends the policy learning model Octo~\cite{team2024octo} by introducing multiple heads, with each head responsible for a specific skill, along with a progress estimator that tracks the progress of skill execution. A skill selector function, named \progss, maps the progress across all skill heads to determine the appropriate skill to execute at any given state. Both the skill heads and progress estimators are trained simultaneously using the same pre-trained Octo transformer backbone. With the multi-head structure, \ours allows for the training of multiple skills either synchronously or asynchronously, facilitating the integration of a large skill set and the addition of new skills as needed.

The main advantage of \ours is that it provides a clear understanding of each individual skill and the overall task progress through continuous progress estimation. It also offers flexibility in determining when to terminate a skill by setting a termination threshold. Additionally, \ours enhances reliability under disturbance, as the model continuously reasons when to skip or redo certain skills to ensure task completion.

Comprehensive experiments in both simulated and real-world environments demonstrate that \ours effectively addresses challenging long-horizon dexterous manipulation tasks, showing significant improvements over the Octo baseline models. In tasks involving flipping, picking, packing, and pushing, \ours increased the overall task completion rate from 32.5\% with the baseline Octo single policy to 90\% with \ours.


\section{Related Works}\label{sec:related}
{\bf Policy Learning} 
Dexterous manipulation is a crucial capability for robots, enabling them to perform complex tasks in the human environment\cite{shah2021rrl,padalkar2023open, ahn2024autort,yang2023moma,zhao2023learning}.
Extensive research has focused on learning dexterous skills through policy learning, both in reinforcement learning and imitation learning, with various input modalities, including vision sensory inputs\cite{andrychowicz2020learning,yang2023moma}, robot states\cite{shah2021rrl,chi2023diffusion,zhao2023learning,team2024octo,mandlekar2021matters,liu2024moka} and language prompts\cite{team2024octo,jiang2022vima}. 
To achieve more flexible goal-oriented tasks, prior works have also explored skill learning conditioned on goal images\cite{team2024octo,jiang2022vima,du2024learning}. 
For example, Du et. al.\cite{du2024learning} generate key image frames to guide robot execution.

Recent advances have involved fine-tuning foundation models pre-trained on large image datasets\cite{shah2021rrl,yang2023moma,chi2023diffusion,xu2023joint} or trajectories from various robot embodiments\cite{padalkar2023open,ahn2024autort,team2024octo}. 
Specifically, Octo model team\cite{team2024octo} pre-trains a transformer-based generalist robot policy with over 800k robot episodes from Open X-Embodiment dataset\cite{open_x_embodiment_rt_x_2023}.
In our work, we explore multiple input modalities, including goal state indicators in natural language and goal images. We fine-tune the Octo transformer to learn a diverse set of skills concurrently.

{\bf Long-horizon manipulation planning} 
Long-horizon manipulation tasks often involve a sequence of sub-tasks, requiring both the selection of the correct motion primitives and the precise execution of each action at every time step. 
Extensive research has focused on sequencing prehensile actions (e.g., pick and place), addressing the combinatorial challenges\cite{garrett2020pddlstream,gao2023minimizing,wang2021uniform} and the need for efficient state sampling\cite{garrett2020pddlstream,chang2023lgmcts,gao2022fast}. However, when non-prehensile actions (e.g., pushing or poking) are introduced, the increased uncertainty in the resulting states after actions complicates long-horizon planning. 
Researchers have tackled this challenge using Monte Carlo tree search\cite{chang2023lgmcts,vieira2023effective}, probabilistic models\cite{mishra2023generative}, and deep reinforcement learning models\cite{nasiriany2022augmenting,agia2023stap}. 
Additionally, previous works have generated sub-goals in the form of robot poses\cite{liu2024moka, belkhale2023hydra} or future camera observations\cite{du2024learning,mandlekar2020learning} to guide robot execution. 
When a manipulation task requires a heterogeneous skill set, a common approach is to leverage a high-level classifier based skill selector to determine which skill to execute. 
Nasiriany et. al.\cite{nasiriany2022augmenting} train a task policy to determine the motion primative to apply. 
Belkhale et. al. \cite{belkhale2023hydra} train a mode head to choose whether moving directly to a distant waypoint or following a dense action sequence.
However, the one-hot classification result provides limited information of the progress in execution and is ambiguously annotated during the transition of skills.
In this work, we propose \progss, a ``skill progress'' guided skill selector to determine the appropriate skill to execute in long-horizon dexterous manipulation tasks. 
Instead of training a skill selector from scratch independent to the skill set, our skill selector shares the transformer backbone with the skills to enhance efficiency in training and inference.

\section{Problem Formulation}\label{sec:prob}
Let the long-horizon task be represented as a sequence of \emph{skills}, denoted by a finite set \( \mathcal{S} = \{s_1, s_2, \dots, s_N\} \), where \( N \) is the total number of skills. The task execution is governed by a policy \( \pi \), which, at each timestep \( t \), takes as input the observation \( o_t \in \mathcal{O} \) and the current robot state \( x_t \in \mathcal{X} \), and outputs the action \( P_t \in \mathcal{A} \), which will be further detailed below for our problem setting.

\subsection{Single Policy Learning}

The observation \( o_t \) is a representation of the environment at time \( t \), which could include sensor readings or environmental context. The robot state \( x_t \) includes the joint configurations, velocities, and other internal variables defining the robot's configuration.

The policy \( \pi \) learns to map:
\[
\pi : (o_t, x_t) \to P_t
\]
where \( P_t \in \mathcal{A} \) is the predicted action. This action consists of two components: the robot’s 6D pose \( p_t = (p_t^x, p_t^y, p_t^z, \theta_t^x, \theta_t^y, \theta_t^z) \in \mathbb{R}^6 \) in Cartesian space and orientation, and a discrete value \( u_t \in \{-1, 0 , 1\} \) representing whether the suction is turned on (\( u_t = 1 \)), turned off (\( u_t = -1 \)), or remains in its current state (\( u_t = 0 \)).

Thus, the action \( P_t \) predicted by the policy can be expressed as:
$P_t = (p_t, u_t)$, 
where \( p_t \) is the 6D pose and \( u_t \) is the suction indicator.

\subsection{\ours: Decomposition of Skills and Progress}

Unlike learning the long-horizon task in one policy, our method, \ours, decomposes the task into skill-specific predictions. For each skill \( s_i \in \mathcal{S} \), we predict not only the robot’s action \( P_t^{(i)} \in \mathcal{A} \), but also a progress value \( \rho_t^{(i)} \in [0, 1] \) that indicates how much of the skill \( s_i \) has been completed at time \( t \). The progress value evolves over time and reaches \( \rho_t^{(i)} = \theta_i \), where \( \theta_i \) is the termination threshold when a skill is considered fully executed.

Thus, for each skill \( s_i \), MuST outputs a tuple:
$(P_t^{(i)}, \rho_t^{(i)})$
at each timestep \( t \), where \( P_t^{(i)} = (p_t^{(i)}, u_t^{(i)}) \) is the predicted action for skill \( s_i \), and \( \rho_t^{(i)} \) is the progress indicator.

\subsection{Skill Selection and Execution (\progss)}

At each timestep \( t \), a skill selector function \( \sigma \) determines the next skill to execute based on the progress values \( \rho_t^{(i)} \) for all skills:
\[
\sigma: \{\rho_t^{(1)}, \rho_t^{(2)}, \dots, \rho_t^{(N)}\} \to s_j
\]
where \( s_j \in \mathcal{S} \) is the selected skill to be executed at time \( t \).

The robot then executes the action \( P_t^{(j)} \) predicted by MuST for the selected skill \( s_j \):
$a_t = P_t^{(j)}$,
where \( a_t \in \mathcal{A} \) represents the robot’s pose and suction status to execute at timestep \( t \).

\subsection{Goal-Conditioned Policy Learning}

In both the single policy learning and MuST approaches, an alternative model variant can be used where the input also encodes a goal condition. This goal condition can be represented as either a goal image \( I_g \in \mathcal{I} \) or a goal language instruction \( l_g \in \mathcal{L} \), where \( \mathcal{I} \) is the set of possible goal images and \( \mathcal{L} \) is the set of possible language instructions.

The policy can then be learned as a mapping that includes the goal condition:
\[
\pi : (o_t, x_t, I_g \text{ or } l_g) \to P_t
\]
where the goal condition, whether in the form of a goal image \( I_g \) or a language instruction \( l_g \), provides additional information to the policy for determining the action \( P_t \).

In this variant, the policy outputs \( P_t = (p_t, u_t) \), as described earlier, but the decision process is now conditioned on the provided goal.


\section{Methodology}\label{sec:method}
\begin{figure}[ht]
    \centering
    \includegraphics[width=0.5\textwidth]{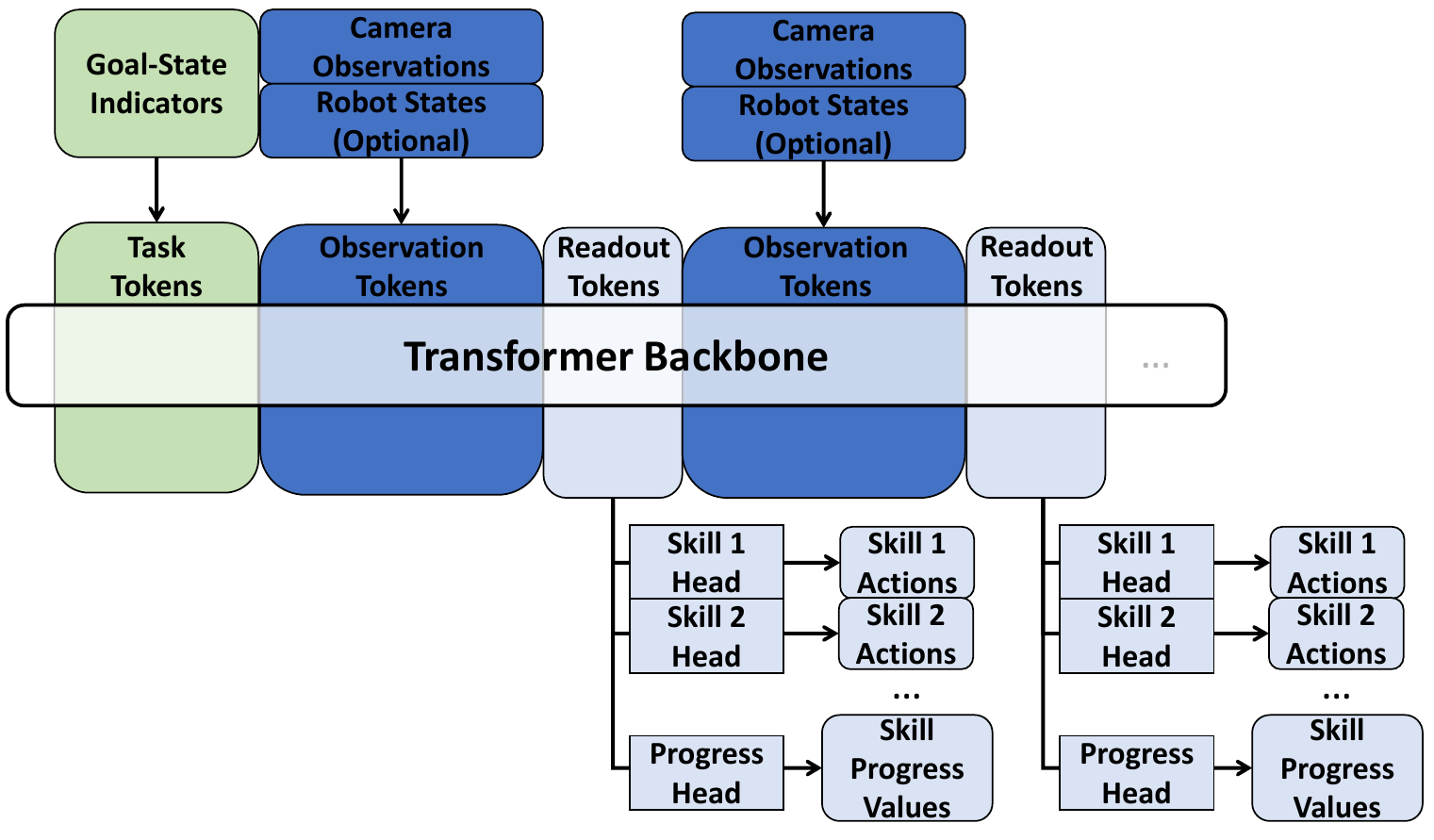}
    \caption{Overview of \ours(Multi-Head Skill Transformer). 
    The model consists of a pre-trained Octo transformer backbone\cite{team2024octo} and $N+1$ heads for an $N-$skill set. 
    Each of the skill head computes an action sequence of its skill.The progress head \progss, the skill selector, estimates the progress of the entire skill set.}
    \label{fig:overview}
\end{figure}
\subsection{Multi-Head Skill Transformer (\ours)}

\ours (Multi-Head Skill Transformer) reduces action complexity by decomposing long-horizon tasks into a set of skills. \ours learns individual skills and then combines them sequentially to complete the overall task.
\ours has $N+1$ action heads, with one action head for each skill in $\mathcal{S}$ and one \emph{progress head} to estimate progress of the skill execution. 
The $N+1$ heads share the same Octo transformer backbone~\cite{team2024octo}, which is pre-trained with 800k robot manipulation episodes.

Fig.~\ref{fig:overview} is \ours architecture overview with three key components: input tokenizers, a transformer backbone, and output decoding heads for N skills, along with a progress head, all extending from the readout tokens.
The input tokenizers and the transformer backbone are the same as in Octo, while the readout token serves as a fused representation of the multi-modal observations.
Taking the readout token as input, \ours extends skill heads and progress heads where each head is an L1 action head which consists of a multi-head attention pooling block followed by a dense layer. Each skill head computes an action for the corresponding skill $P_t^{(i)}, \text{ for all } i = 1, 2, \dots, N$, 
 and the progress head learns a numerical vector $\{\rho_t^{(1)}, \rho_t^{(2)}, \dots, \rho_t^{(N)}\}$ to estimate the current progress of each skill. 

\subsection{\progss: Progress Guided Skill Selector}\label{sec:progss}

Given the progress vector from the progress head, we design a progress-guided skill selector, \progss, to determine the optimal skill to execute at the current state. Progress values range from 0 to 1, where lower values indicate pending or ongoing execution, and higher values suggest completion. These values help identify the current phase of execution in long-horizon manipulation.

\subsubsection{Object-Centric Progress Annotation}
We define each skill's trajectory segment to consist of three parts: the pre-skill transit trajectory, the skill execution trajectory, and the post-skill transit trajectory.

Skill execution period is defined as the time between the robot's first and last contact with the object. We increase progress value only during this phase, making the progress value object-centric and less sensitive to the robot pose.

\begin{wrapfigure}{r}{0.25\textwidth}  
    \centering
    \includegraphics[width=0.25\textwidth]{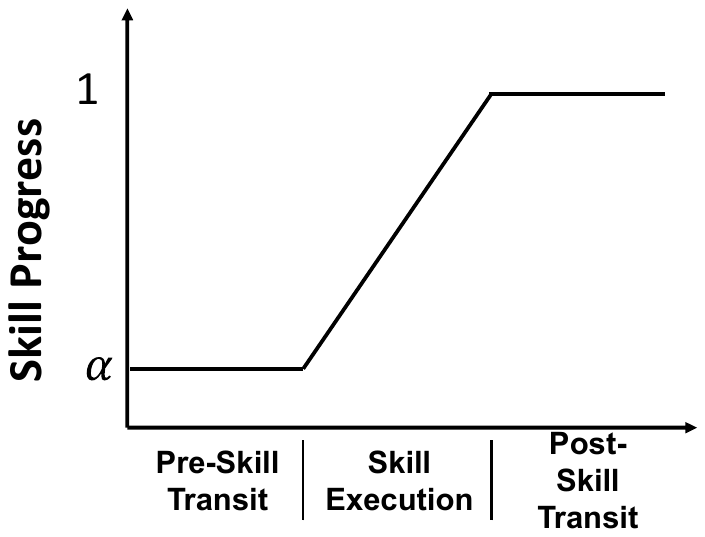}
    \vspace{-7mm}
    \caption{Annotation of skill progress in a skill-related episode segment.}
    \label{fig:annotation}
\end{wrapfigure}

Fig.~\ref{fig:annotation} shows the skills progress annotation of a skill-related trajectory segmentation, for pre- and post-skill transit trajectories, we set the progress value to be $\alpha$ and $1$ respectively. $\alpha:=1-t/M$, where $t$ is the duration of skill execution in the current episode and $M$ is the maximum duration of skill execution among the demonstration episodes of the skill.
The progress value increases linearly with the time steps in demonstration episodes. 

\subsubsection{Skill Selection with Single Skill Sequence}
\label{sec:sequential_progss}
\begin{figure}[H]
    \centering
    \includegraphics[width=0.5\textwidth]{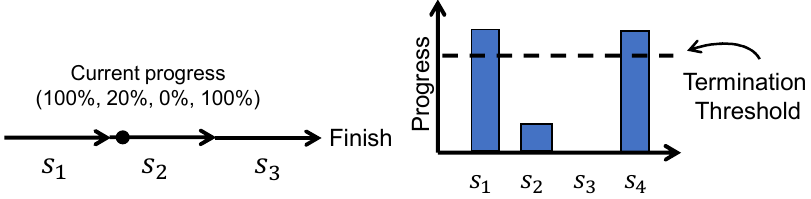}
    \vspace{-7mm}
    \caption{An example of \progss with a single skill sequence.}
    \label{fig:exp-skill-selector}
\end{figure}

When only one skill sequence $\tau=\{s_i\}_{i=1}^{|\tau|}$ of the manipulation task is demonstrated, \progss selects the first skill $s_i$ in the sequence with the current value $\rho^{(i)}_t $ below its termination threshold $\theta_i$. 
Fig.~\ref{fig:exp-skill-selector} is an example of three skills in one sequence $s_1\rightarrow s_2\rightarrow s_3$. 
When \progss with 4 skills infers a vector of $(1, 0.2, 0, 1)$, $s_2$ is selected for execution next for $\theta_i = 0.9, \forall 1\leq i \leq 4$. A side note that $s_4$ is not in the skill sequence, thus its progress value is always $1$.
\vspace{-1mm}
%

When a skill \( s_i \) is divided into segments (\( k > 1 \)), the progress value \( \rho^{(i)}_t \) increments in steps as the robot completes each segment. Instead of increasing continuously from \( \alpha \) to 1, the progress moves between stages, each representing a segment of the skill. For each segment \( j \), progress \( \rho^{(i)}_t \) increases from \( \alpha^i_j \) (start of the segment) to \( \alpha^i_{j+1} \) (end of the segment). The upper bound progress for segment \( j \) is calculated as:
\[
\alpha^i_j = \alpha^i_{j-1} + \frac{(1 - \alpha) \cdot T_j^i}{\sum_{j=1}^k T_j^i}
\]
In this formula, \( \alpha^i_0 = \alpha \) is the initial value. \( T_j^i \) is the duration of segment \( j \) of skill \( s_i \), and \( \sum_{j=1}^k T_j^i \) is the total duration of all segments for skill \( s_i \). \( (1 - \alpha) \) represents the available progress range (from \( \alpha \) to 1). Thus, the progress value increases based on the relative duration of each segment.

At time \( t \), \progss selects the first segment \( j \) of skill \( s_i \) where the progress \( \rho^{(i)}_t \) is below both the termination threshold \( \theta_i \) and the upper bound \( \alpha^i_j \), i.e. 
\(
\rho^{(i)}_t < \min(\theta_i, \alpha^i_j)
\).

Here, \( \theta_i \) is the termination threshold for skill \( s_i \), and \( \alpha^i_j \) is the progress upper bound for segment \( j \). This ensures that the robot progresses through the skill in an incremental manner, selecting the next segment that is not yet completed.

\subsubsection{Skill Selection with Multiple Skill Sequences}
When multiple skill sequences $\{\tau_i\}$ are demonstrated for any combination of skills, Alg.~\ref{alg:arbitrary_progss} is used to select next skill.  
We compute the progress trajectories of demonstrated skill sequences, resulting in a map, $M$, in the skill progress space (line 1).   
\progss searches for the nearest progress trajectory of $\rho_t$ in $M$ and outputs the corresponding skill sequence $\tau$ (Line 2).
We choose the first skill $s_i\in \tau$ with progress value below its threshold, i.e., $\rho^{(i)}_t<\theta_i$ (Line 3-4).

\begin{algorithm}
\begin{small}
    \SetKwInOut{Input}{Input}
    \SetKwInOut{Output}{Output}
    \SetKwComment{Comment}{\% }{}
    \caption{Arbitrary-Sequence-Skill-Selector}
		\label{alg:arbitrary_progss}
    \SetAlgoLined
		\vspace{0.5mm}
    \Input{$T$: Demonstrated skill orderings;\\ 
    $\rho_t$: Estimated progress values;\\
    $\Theta$: Progress termination thresholds.
    }
    \Output{$s$: next skill to execute.}
\vspace{0.5mm}
$M\leftarrow$ Progress-Trajectory-Computation($T$)\\
$\tau\leftarrow$ Nearest-Demonstrations($\rho_t,T,M$).\\
\For{$s_i \in \tau$}{\lIf{$\rho^{(i)}_t<\theta_i$}{\Return $s_i$}}
\end{small}
\end{algorithm}


\subsection{Skill Set Training and Expansion}






\ours multi-head architecture enables concurrent training for a skill set, fine-tuning of a single skill with additional data, and expansion of the existing skill set. For the batch training, we alternate training of  $s_i$ in the set, updating the parameters of the transformer backbone, its skill head, and the progress head. We freeze the parameters of the transformer backbone for fine-tuning with additional data or expansion. For new skill to be introduced into the existing model, we add a new action head and extend the dimension of progress head, then train the new skill head and the progress head. The performance of other skills is not affected during fine-tuning or expansion with frozen backbone.

\section{Evaluation}\label{sec:experiments}
In this section, we evaluate the performance of \ours in both simulated and real-robot environments, with a total of six experiments, under the various conditions:
\begin{itemize}
    \item Performance gain of MuST in comparison with the single-head Octo model 
    
   \item  Performance of MuST, conditioned on task goals specified by images or language instructions
    
    \item Performance of MuST across a diverse set of objects
    
    \item Ability for MuST to react to unexpected environment disturbance, based on the skill progress values
\end{itemize}

The performance is measured by two  metrics:
\begin{enumerate}
    \item {\bf Skill completion and task completion:} we report the completion of a single skill, and the task is completed only if all skills in the task have been successful executed. Any failed skill will result in failure of all subsequent skills.
    \item {\bf Execution time:} The execution time is defined as spent time for manipulation until the object consistently stays at the goal pose $g$ for 100 time steps. 
\end{enumerate}

We compare \ours with the single-head Octo model, which learn all skills without progress estimation.
Both models finetune Octo-Base (93M params) checkpoint and use L1 action heads as decoding heads for skills and progress. 
For closed-loop control scenarios, both models carry out inference every 50 time steps and compute the action sequence for the next 50 time steps.
We train the models with an Nvidia V100 16GB GPU.

\subsection{Simulation Experimental Results and Comparison with Octo Baseline  }

Our goal-state conditioned Pick-n-Pack manipulation task Fig.~\ref{fig:intro}[Top] consists of four skills:
First, the robot flips down the object from the tote boundary to enable picking. 
Next, it picks the object from the picking tote.
Based on the goal-state indicator $I_g$ or $l_g$, the robot packs the object near the desired corner of the packing tote.
Finally, the robot pushs the object, both  rotating and translating it, to fit it tightly the desired corner.

\begin{figure}
    \centering
    \includegraphics[width=0.5\textwidth]{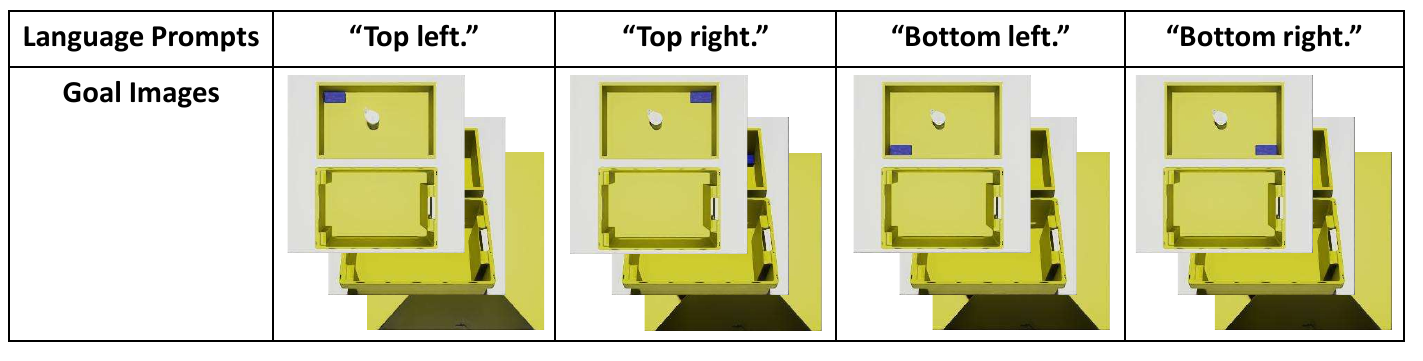}
    \caption{We use either language prompts or images as goal state indicators to customize packing poses.}
    \label{fig:end_state_indicator}
\end{figure}

The goal is given by either a language prompt or a goal image Fig.~\ref{fig:end_state_indicator} . 
For the goal images, instead of specific images of objects, we use a blue patch to suggest the goal state which makes it independent of object appearance, enhancing the model's generalization capability across different object types.

\begin{figure}
    \centering
    \includegraphics[width=0.37\textwidth]{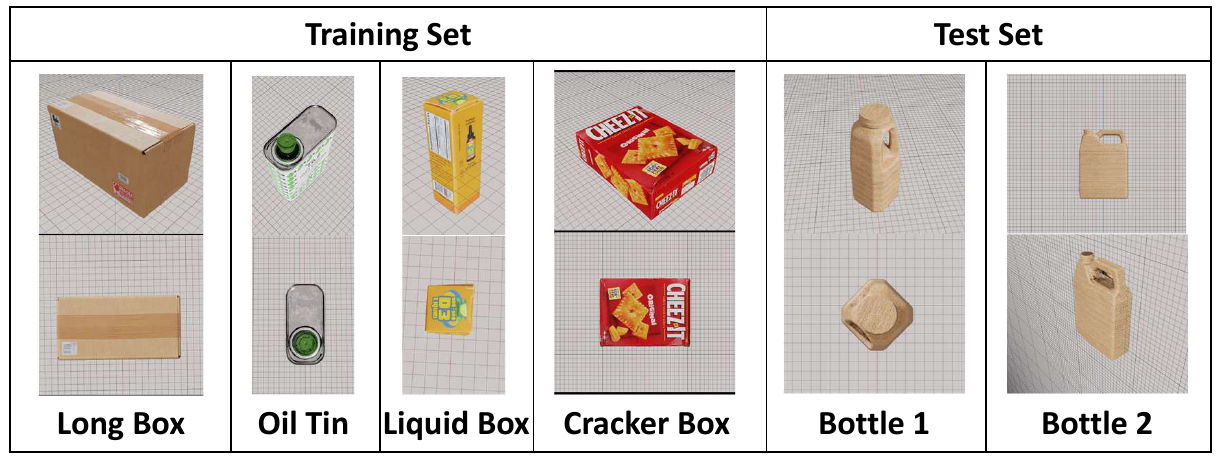}
    \caption{Training object set and test object set in simulation. The 3D model used are open-source models sampled from the YCB Object and Model Set~\cite{YCB}, NVIDIA SimReady assets~\cite{nvidia_simready}, open-source models from SketchFab~\cite{sketchfab_open_source}, and the Google Scanned Objects dataset~\cite{2022googlescannedobjectshighquality}.}
    \label{fig:object_set}
\end{figure}

Tab.\ref{tab:single_language_results} reports the performance metrics for both  \ours  and the Single Head Octo. The task is language conditioned Pick-n-Pack with the ``long box''(Fig.~\ref{fig:object_set}), and each task is repeated 10 times. While Octo model succeeded in  $32.5\%$ tasks, \ours maintains $80\%-90\%$ success rate.
In addition, \ours is $23.7\%-38.4\%$ faster than Octo in execution time for the finished tasks.
The results suggest that \ours is more robust than Octo baseline in long horizon manipulation tasks. 

\scriptsize
\begin{table}[ht]
\centering
\caption{Language Conditioned Pick-n-Pack (Long Box)}
\label{tab:single_language_results}
\begin{tabular}{@{\centering\arraybackslash}m{1.4cm}>{\centering\arraybackslash}m{0.8cm}>{\centering\arraybackslash}m{0.8cm}>{\centering\arraybackslash}m{0.8cm}>{\centering\arraybackslash}m{0.8cm}>
{\centering\arraybackslash}m{0.8cm}>{\centering\arraybackslash}m{0.8cm}>{\centering\arraybackslash}m{0.8cm}>{\centering\arraybackslash}m{0.8cm}>{\centering\arraybackslash}m{0.8cm}@{}}
\toprule
 & \multicolumn{6}{c}{\textbf{Task Completion}} \\ \cmidrule(lr){2-7} 
 & \multicolumn{2}{c}{\scriptsize{\textbf{Flip$\rightarrow$Pack}}} & \multicolumn{2}{c}{\textbf{{\scriptsize Push (Orientation)}}} & \multicolumn{2}{c}{{\scriptsize \textbf{Push \newline (Position)}}} & \multicolumn{2}{c}{{\scriptsize \textbf{Execution \newline Time}}}\\ \cmidrule(lr){2-3} \cmidrule(lr){4-5} \cmidrule(lr){6-7} \cmidrule(lr){8-9}
 \textbf{End State} & \textbf{\ours} & \textbf{Octo} & \textbf{\ours} & \textbf{Octo}& \textbf{\ours} & \textbf{Octo}& \textbf{\ours} & \textbf{Octo} \\ \midrule
{\scriptsize Top Left} & \multicolumn{1}{|c}{10/10} & 7/10 & \multicolumn{1}{|c}{10/10} & 4/10 & \multicolumn{1}{|c}{10/10} & 4/10 & \multicolumn{1}{|c}{1324} & 1734 \\
{\scriptsize Top Right} & \multicolumn{1}{|c}{9/10} & 9/10 & \multicolumn{1}{|c}{8/10} & 7/10 & \multicolumn{1}{|c}{8/10} & 4/10 & \multicolumn{1}{|c}{1530} & 2639 \\
{\scriptsize Bottom Left} & \multicolumn{1}{|c}{10/10} & 7/10 & \multicolumn{1}{|c}{9/10} & 5/10 & \multicolumn{1}{|c}{9/10} & 3/10 & \multicolumn{1}{|c}{1538} & 2337\\
{\scriptsize Bottom Right} & \multicolumn{1}{|c}{9/10} & 6/10 & \multicolumn{1}{|c}{9/10} & 4/10 & \multicolumn{1}{|c}{9/10} & 2/10 & \multicolumn{1}{|c}{1571} & 2550 \\ \bottomrule
\end{tabular}
\end{table}
\normalsize

\subsection{Additional Evaluation of \ours and \progss in Simulation}
In this section, we further evaluate the performance of \ours, conditioned with goal images, on a diverse object set, and on progress estimation.

Tab.~\ref{tab:single_image_results} summarizes  the performance of \ours on image-conditioned Pick-n-Pack task on the same Long Box object. \ours maintains $80\%-90\%$ success rate for 40 evaluation trials.

Furthermore, we evaluate \ours on a diverse object set. In this experiment, we train \ours on four different boxes(Fig.~\ref{fig:object_set}). The training dataset include 15 demonstrations for each of training objects at each of the four goal poses.
We test \ours on all objects including four training objects and two novel objects, and results are reported in Tab.~\ref{tab:four_obj_language_results} for 20 trials on each object.
For the first three skills, \ours maintains over $90\%$ completion rate on training object set and that drops to $70\%-75\%$ on objects in the new category.
For the last push skill, \ours solves $80\%$ test cases on the cracker box, $65\%-70\%$ on other training objects.
\ours only finishes around $40\%$ pushes on novel objects. 
In the training dataset with cuboid objects, the push demonstrations use box corners for rotation.  Our hypothesis is that the same push behavior does not generalize well to the tested novel objects with irregular shapes.

\begin{table}[H]
\centering
\caption{Image Conditioned Pick-n-Pack (Long Box)}
\label{tab:single_image_results}
\begin{tabular}{@{\centering\arraybackslash}m{1.7cm}>{\centering\arraybackslash}m{1cm}>{\centering\arraybackslash}m{1cm}>{\centering\arraybackslash}m{1cm}>{\centering\arraybackslash}m{1.5cm}>{\centering\arraybackslash}m{1.2cm}>{\centering\arraybackslash}m{1cm}@{}}
\toprule
 & \multicolumn{5}{c}{\textbf{Task Completion}} \\ \cmidrule(lr){2-6} 
\textbf{Packing \newline Corner} & \textbf{Flip} & \textbf{Pick} & \textbf{Pack} & \textbf{Push (Orientation)} & \textbf{Push (Position)} & \textbf{Execution Time}\\ \midrule
Top Left & 10/10 & 10/10 & 10/10 & 9/10 & 8/10 & 1570 \\
Top Right & 9/10 & 9/10 & 9/10 & 8/10 & 8/10 & 1765 \\
Bottom Left & 10/10 & 10/10 & 10/10 & 9/10 & 9/10 & 1225 \\
Bottom Right & 10/10 & 10/10 & 10/10 & 8/10 & 9/10 & 1158 \\ \bottomrule
\end{tabular}
\end{table}

\begin{table}[ht]
\centering
\caption{Language-Conditioned Pick-n-Pack with Diverse Object Set}
\label{tab:four_obj_language_results}
\begin{tabular}{@{\centering\arraybackslash}m{1.7cm}>{\centering\arraybackslash}m{1cm}>{\centering\arraybackslash}m{1cm}>{\centering\arraybackslash}m{1cm}>{\centering\arraybackslash}m{1.5cm}>{\centering\arraybackslash}m{1.2cm}>{\centering\arraybackslash}m{1cm}@{}}
\toprule
 & \multicolumn{5}{c}{\textbf{Task Completion}} \\ \cmidrule(lr){2-6} 
\textbf{Test Object} & \textbf{Flip} & \textbf{Pick} & \textbf{Pack} & \textbf{Push (Orientation)} & \textbf{Push (Position)} & \textbf{Execution Time}\\ \midrule
Cracker Box & 20/20 & 20/20 & 19/20 & 16/20 & 16/20 & 2073 \\
Liquid Box & 20/20 & 20/20 & 19/20 & 14/20 & 13/20 & 1769 \\
Long Box & 20/20 & 20/20 & 18/20 & 14/20 & 13/20 & 1330 \\
Oil Tin & 20/20 & 20/20 & 19/20 & 14/20 & 13/20 & 1875 \\ \midrule
Bottle1 (OOD) & 20/20 & 20/20 & 15/20 & 12/20 & 9/20 & 1608\\ 
Bottle2 (OOD) & 20/20 & 20/20 & 14/20 & 9/20 & 7/20 & 2009\\ \bottomrule
\end{tabular}
\end{table}

Additionally, we demonstrate that \ours can react to unexpected environment disturbance based on the skill progress values. As an example, when user resets an object on the tote edge, \ours would select the Flip skill repeatedly. Similarly, \ours would skip the first skill, and start with the second Pick skill when the object is in the pick state. We include these demonstrations with the skill progress value graphs in the accompanying video.

\subsection{Handling Multiple Sequences in Simulation}\label{sec:sim_task2}

\begin{figure}[ht]
    \centering
    \includegraphics[width=0.35\textwidth]{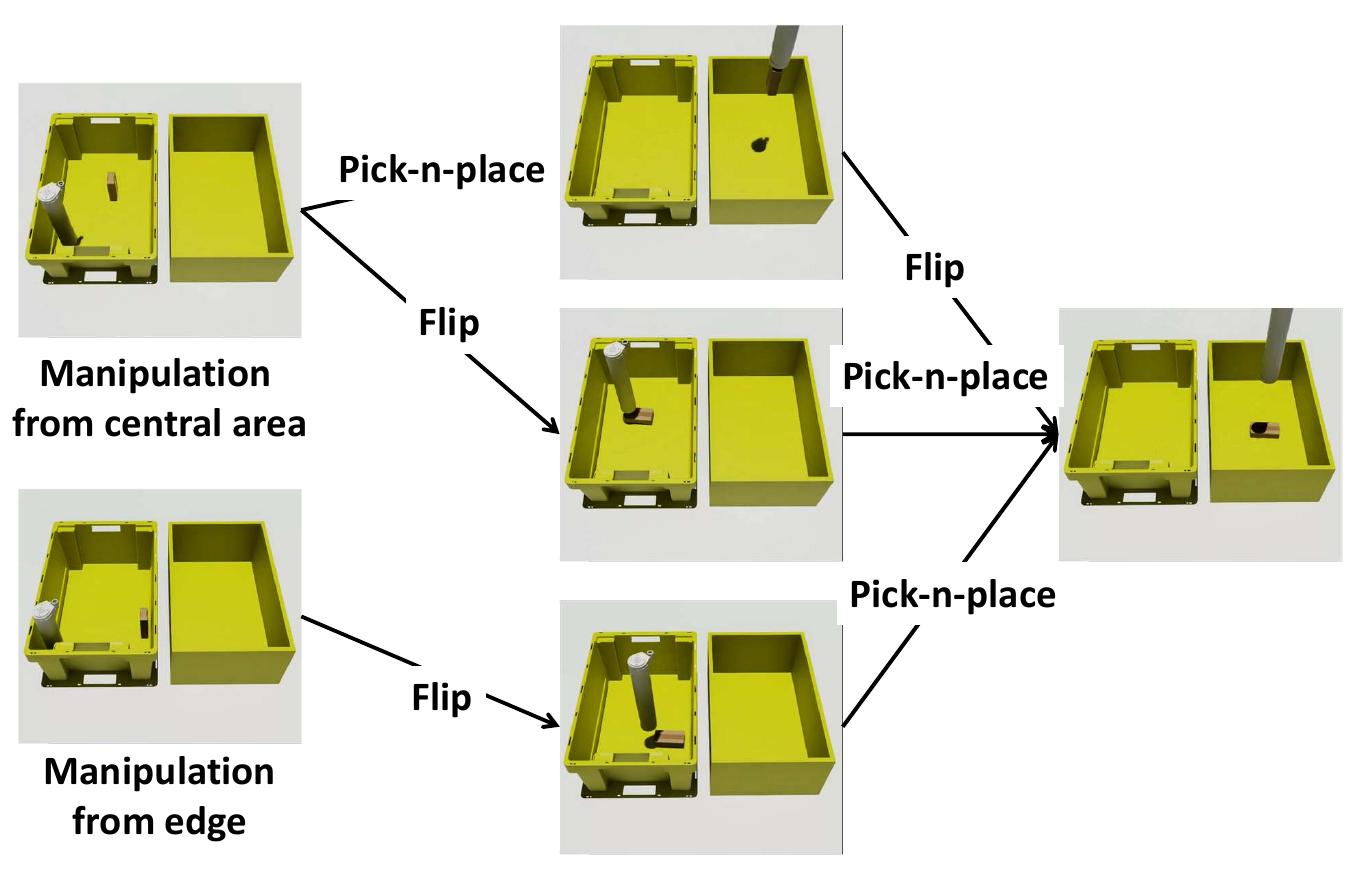}
    \vspace{-3mm}
    \caption{Multi-sequence pick-n-pack. When the object is sampled at the central area of the tote, there are two possible skill orderings; when the object is sampled at the edge, the robot cannot directly pick it up before flipping.}
    \label{fig:multi_sequence}
\end{figure}

We designed the task of multi-sequence pick-n-pack to evaluate the performance of \ours when multiple skill sequences are demonstrated.
As shown in Fig.~\ref{fig:multi_sequence}, the robot is tasked to flip down a box and place it at the packing tote. 
Specifically, when the object is located in the central area of the tote, the robot can choose to flip the object before or after pick-n-place.
However, when the object is located at the boundary of the picking tote, the robot cannot execute pick-n-place before a successful flip.
For each of the three cases of skill sequences in Fig.~\ref{fig:multi_sequence}, we made 50 demonstrations. 

Tab.~\ref{tab:multi_sequence_results} shows sequence selection distribution and task completion rate.
When the object is sampled at the central area of the picking tote, out of the 80 trials, \ours chooses ``pick first'' and ``flip first'' sequences in $37.5\%$ and $62.5\%$ trials respectively with around $80\%$ success rate on both skill sequences.
When the object is sampled at the edge of the picking tote, picking directly is impossible. 
\ours chooses the ``flip first'' sequence in $96.2\%$ test cases.
The results suggest that \ours effectively handles multiple skill sequences and avoids ``modality collapses'' in long horizon manipulation tasks.

\begin{table}[H]
\centering
\caption{Multi-Sequence Pick-n-Pack}
\label{tab:multi_sequence_results}
\begin{tabular}{@{\centering\arraybackslash}m{4cm}>{\centering\arraybackslash}m{2cm}>{\centering\arraybackslash}m{2cm}@{}}
\toprule
\textbf{Test Cases} & \textbf{Pick First} & \textbf{Flip First}\\ \midrule
Manipulate from central area & 24/30 & 39/50\\
Manipulate from edge & 0/3 & 73/77\\ \bottomrule
\end{tabular}
\end{table}


\subsection{Physical Experimental Results}\label{sec:real_task}
\begin{wrapfigure}{r}{0.25\textwidth}  
    \centering
    \includegraphics[width=0.25\textwidth]{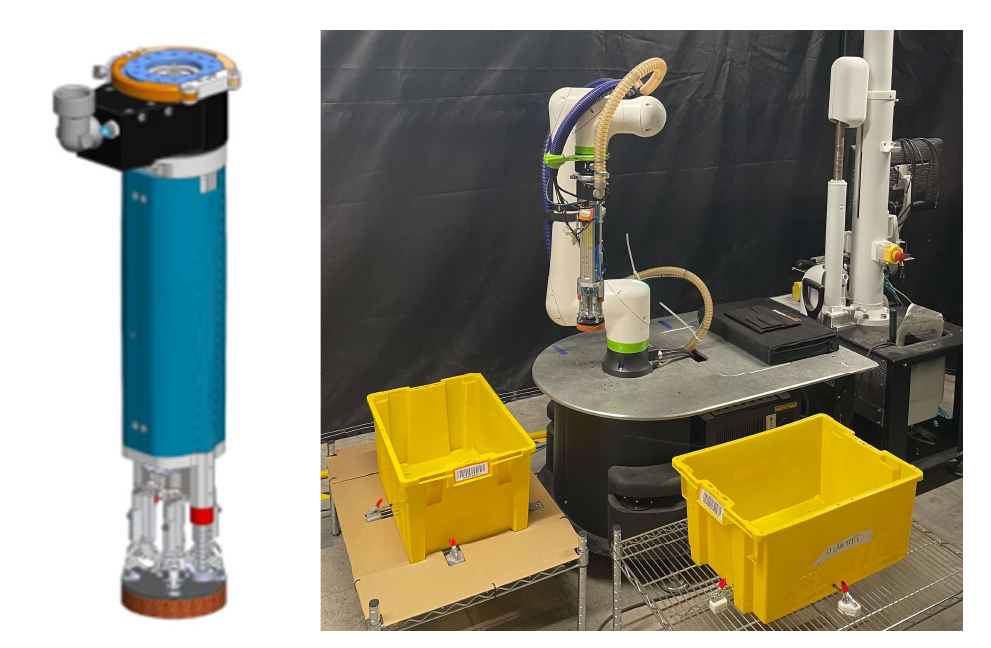}
    \caption{[Left] A customized suction gripper capable of vacuum suction and dexterous contact. [Right] Physical robotic system.}
    \vspace{-1mm}
    \label{fig:sock_puppet}
\end{wrapfigure}

Our robotic test bed (Fig.~\ref{fig:sock_puppet}[Right]) comprises a collaborative manipulator equipped with a customized suction gripper(Fig.~\ref{fig:sock_puppet}[Left]), which is capable of vacuum suction and dexterous contact with its soft tip. Two 5 MP 3D cameras are positioned with one above each tote. 

\begin{figure}[t]
    \centering
    \includegraphics[width=0.4\textwidth]{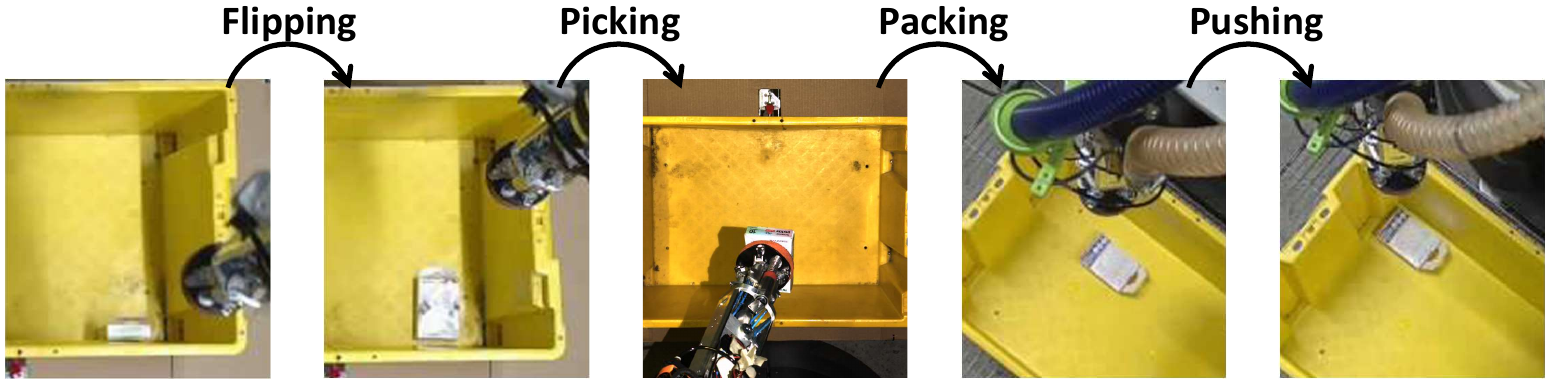}
    \vspace{-2mm}
    \caption{Task sequence of real robot goal-state conditioned pick-n-pack.}
    \label{fig:station2_sequence}
\end{figure}

\begin{figure}
    \centering
    \vspace{-3mm}
    \includegraphics[width=0.4\textwidth]{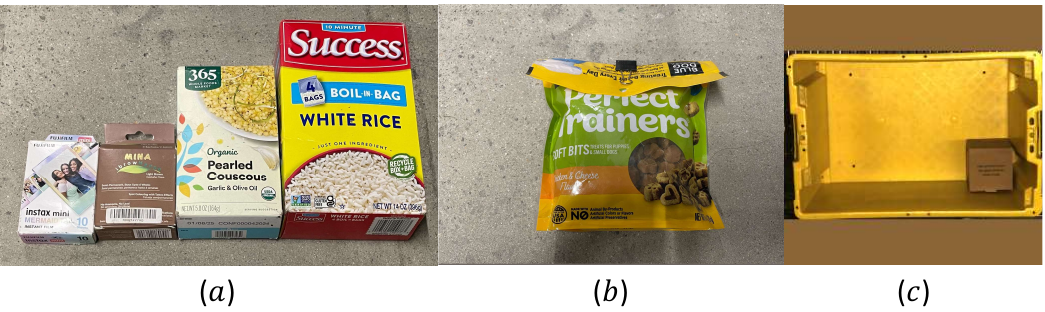}
    \vspace{-3mm}
    \caption{Training (a) and test (b) object set for physical experiments. (c) Image of the packing tote with a universal object as the goal-state indicator.}
    \label{fig:real_images}
\end{figure}

Similarly, the experimental task (Fig.~\ref{fig:station2_sequence}) consists of four skills: flips down an object from the edge of the picking tote, grasps it with the suction cup, packs it at the proper pose, based on a goal image using a generic brown box(Fig.~\ref{fig:real_images}(c), and pushes it to the corners of the tote.
The first two skills have clear success or failure criteria, while packing succeeds if the object is in the correct quarter of the tote, and pushing succeeds if the object is within 2 cm of the correct corner. 

We evaluate \ours in a open-loop control framework, where \ours takes a single state observation, two images of two totes plus an image of the in-hand object (if any), and outputs the trajectory for the selected skill. The robot then executes the   entire trajectory in an open loop and moves out of the observable environment. We use a set of five objects (Fig.~\ref{fig:real_images} for the physical experimental task. For each object in our training set (Fig.~\ref{fig:real_images}(a)), we collect 15, 15, 24, and 24 human demonstrations for the four skills respectively. 

We first evaluate both \ours and Octo on the couscous box with different goal-state images (Tab.~\ref{tab:real_robot_diff_corner_results}).
Both models succeed in the first three skills in all the test cases but Octo only successfully pushes the object to corners in $35\%$ test cases.
In Tab.~\ref{tab:real_robot_diff_obj_results}, we show the experiment results on the diverse object set.
Octo has $0\%$ success rate on the small object (``Instax'' box) and the novel object (Bag).
It also fails in pushing other objects to corners in most test cases.
In contrast, \ours finishes $88\%$ test cases among the objects.
In addition, most of the failures of Octo in pushing are due to collisions with the environment, while two failures of \ours on pushing are inaccurate location with the open loop limitation: the final object pose is slightly outside the goal region (2.1 cm and 2.6 cm away from the corner).
In summary, \ours has much higher success rate in finishing the whole task than Octo, especially on small objects and novel objects.

\begin{table}[H]
\centering
\caption{Four Image Conditioned Goals(Couscous Box)}
\label{tab:real_robot_diff_corner_results}
\begin{tabular}{@{\centering\arraybackslash}m{1.7cm}>{\centering\arraybackslash}m{0.8cm}>{\centering\arraybackslash}m{0.8cm}>{\centering\arraybackslash}m{0.8cm}>{\centering\arraybackslash}m{0.8cm}>{\centering\arraybackslash}m{0.8cm}>{\centering\arraybackslash}m{0.8cm}>{\centering\arraybackslash}m{0.8cm}>{\centering\arraybackslash}m{1cm}@{}}
\toprule
 & \multicolumn{2}{c}{\textbf{Flip}} & \multicolumn{2}{c}{\textbf{Pick}} & \multicolumn{2}{c}{\textbf{Pack}} & \multicolumn{2}{c}{\textbf{Push}} \\ \cmidrule(lr){2-3} \cmidrule(lr){4-5} \cmidrule(lr){6-7}  \cmidrule(lr){8-9}
 \textbf{Goal State} & \textbf{\ours} & \textbf{Octo} & \textbf{\ours} & \textbf{Octo} & \textbf{\ours} & \textbf{Octo} & \textbf{\ours} & \textbf{Octo} \\ \midrule
Top Left & \multicolumn{1}{|c}{5/5} & 5/5 & \multicolumn{1}{|c}{5/5} & 5/5 & \multicolumn{1}{|c}{5/5} & 5/5 & \multicolumn{1}{|c}{4/5} & 3/5 \\
Top Right & \multicolumn{1}{|c}{5/5} & 5/5 & \multicolumn{1}{|c}{5/5} & 5/5 & \multicolumn{1}{|c}{5/5} & 5/5 & \multicolumn{1}{|c}{2/5} & 0/5 \\
Bottom Left & \multicolumn{1}{|c}{5/5} & 5/5 & \multicolumn{1}{|c}{5/5} & 5/5 & \multicolumn{1}{|c}{5/5} & 5/5 & \multicolumn{1}{|c}{5/5} & 2/5 \\
Bottom Right & \multicolumn{1}{|c}{5/5} & 5/5 & \multicolumn{1}{|c}{5/5} & 5/5 & \multicolumn{1}{|c}{5/5} & 5/5 & \multicolumn{1}{|c}{5/5} & 2/5 \\ 
 \bottomrule
\end{tabular}
\end{table}

\begin{table}[H]
\centering
\caption{Five Objects Image-Conditioned (Bottom Left Corner)}
\label{tab:real_robot_diff_obj_results}
\begin{tabular}{@{\centering\arraybackslash}m{1.7cm}
>{\centering\arraybackslash}m{0.8cm}
>{\centering\arraybackslash}m{0.8cm}
>{\centering\arraybackslash}m{0.8cm}
>{\centering\arraybackslash}m{0.8cm}
>{\centering\arraybackslash}m{0.8cm}
>{\centering\arraybackslash}m{0.8cm}
>{\centering\arraybackslash}m{0.8cm}
>{\centering\arraybackslash}m{0.8cm}@{}}
\toprule
 & \multicolumn{2}{c}{\textbf{Flip}} & \multicolumn{2}{c}{\textbf{Pick}} & \multicolumn{2}{c}{\textbf{Pack}} & \multicolumn{2}{c}{\textbf{Push}} \\ \cmidrule(lr){2-3} \cmidrule(lr){4-5} \cmidrule(lr){6-7}  \cmidrule(lr){8-9}
 \textbf{Object} & \textbf{\ours} & \textbf{Octo} & \textbf{\ours} & \textbf{Octo} & \textbf{\ours} & \textbf{Octo} & \textbf{\ours} & \textbf{Octo} \\ \midrule
``Instax'' Box & \multicolumn{1}{|c}{ 4/5} & 0/5 & \multicolumn{1}{|c}{4/5} & 0/5 & \multicolumn{1}{|c}{4/5} & 0/5 & \multicolumn{1}{|c}{2/5} & 0/5 \\
``Mina'' Box & \multicolumn{1}{|c}{5/5} & 4/5 & \multicolumn{1}{|c}{5/5} & 4/5 & \multicolumn{1}{|c}{5/5} & 4/5 & \multicolumn{1}{|c}{5/5} & 3/5 \\
Couscous Box & \multicolumn{1}{|c}{5/5} & 5/5 & \multicolumn{1}{|c}{5/5} & 5/5 & \multicolumn{1}{|c}{5/5} & 5/5 & \multicolumn{1}{|c}{5/5} & 2/5 \\
Rice Box & \multicolumn{1}{|c}{ 5/5} & 5/5 & \multicolumn{1}{|c}{5/5} & 5/5 & \multicolumn{1}{|c}{5/5} & 4/5 & \multicolumn{1}{|c}{5/5} & 1/5 \\ 
Bag (OOD) &\multicolumn{1}{|c}{ 5/5} & 0/5 & \multicolumn{1}{|c}{5/5} & 0/5 & \multicolumn{1}{|c}{5/5} & 0/5 & \multicolumn{1}{|c}{5/5} & 0/5  \\ \bottomrule
\end{tabular}
\vspace{-3mm}
\end{table}


\section{Conclusion}\label{sec:conclusion}
In this work, we introduce \ours, a skill chaining model designed for long horizon dexterous manipulation. 
Given a task decomposed into $N$ skills, \ours trains $N+1$ heads: $N$ heads to learn individual skills and an additional head for skill progress estimation. 
Based on the estimated progress values, a skill progress guided skill selector \progss chooses the proper skill to execute at each time step. 
Qualitative results demonstrate that \progss effectively adapts to unexpected disturbance.
Comprehensive experiments in both simulation and real world settings reveal the performance advantages of \ours over the single-head Octo baseline, as well as its capability to handle various skill sequences and diverse object sets.

\newpage
\bibliographystyle{format/IEEEtran}
\bibliography{bib/bib}

\end{document}